# Second Order Probabilities for Uncertain and Conflicting Evidence

Gerhard Paaß*
GMD, D-5205 Sankt Augustin
email: paass@gmdzi.gmd.de

## Abstract

In this paper the elicitation of probabilities from human experts is considered as a measurement process, which may be disturbed by random 'measurement noise'. Using Bayesian concepts a second order probability distribution is derived reflecting the uncertainty of the input probabilities. The algorithm is based on an approximate sample representation of the basic probabilities. This sample is continuously modified by a stochastic simulation procedure, the Metropolis algorithm, such that the sequence of successive samples corresponds to the desired posterior distribution. The procedure is able to combine inconsistent probabilities according to their reliability and is applicable to general inference networks with arbitrary structure. Dempster-Shafer probability mass functions may be included using specific measurement distributions. The properties of the approach are demonstrated by numerical experiments.

## 1 Introduction

Most probabilistic inference systems pretend that the available probabilities are exact point values. Psychological studies [Hogarth 1987], however, show that human experts have only moderate abilities as probability assessors and often are unable to specify point estimates. If an expert estimates several interrelated probabilities these often are *incoherent* and violate the laws of probability calculus. In this situation update mechanisms for point probabilities [Pearl 1988; Lauritzen & Spiegelhalter 1988] cannot work as there *exists no* probability measure that simultaneously meets all constraints. The rationality of experts, which is the basis of subjective probability theories [Cheeseman 1988], seems to be only a theoretical ideal and does not apply to practical situations.

We consider the elicitation of probabilities by experts to be some sort of *measurement process* which may be disturbed by 'measurement noise'. This noise again is assumed to be random and therefore induces a second order probability measure, which describes the distribution of the probabilities delivered by the experts. The application of Bayesian principles allows to take into account various forms of information on marginal probabilities, conditional probabilities, probability intervals, and arbitrary other parameters of the probability measure always considering their relative precision. In contrast to probability intervals, which always reflect worst case errors, the utilization of measurement distributions allows to express the relative plausibility of different feasible values. Consequently the derived imprecision of result probabilities in general will be smaller.

In the paper the basic 'first order' probability distribution $p$ is represented approximately by a random sample of 'possible worlds'. Then the desired second order posterior distribution $P(p)$ evolves during the continuous modification of this sample with the Metropolis algorithm, which is the essential component of the simulated annealing algorithm [Mitra & al. 1986]. The underlying inference networks may have a general structure with cycles. This contrasts to causal probabilistic networks, where the nodes

---

*This work was supported by the German Federal Department of Research and Technology, grant ITW8900A7

have a causal ordering [Pearl 1988, Andersen & al. 1989].

Imprecise probabilities are usually represented by probability intervals [Nilsson 1986, Grosof 1986], e.g. $p(A) \in [a, b]$. An uncertain version is the statement: "In 90% of the cases $p(A) \in [a, b]$ holds". The algorithm described below can process such uncertain probability intervals. The utilization of second order probability distributions was first proposed by [Lindley & al. 1979] to pool the opinions of the members of a committee. A discussion is given by [Genest & Zidek 1986, Paass 1988, Kyburg 1989]. Stochastic simulation approaches for probabilistic reasoning have been used by [Pearl 1988, Henrion 1988] to determine 'first order' probability distributions for precise and consistent input probabilities.

## 2 Probabilistic Model

Suppose the situation in question can be specified in terms of $k$ different basic propositions $U_1, \ldots, U_k$. A term $W_\tau := \tilde{U}_1 \wedge \cdots \wedge \tilde{U}_k$ with $\tilde{U}_i \in \{U_i, \neg U_i\}$ embodies a comprehensive description of the situation and is called a possible world. Define $\mathcal{F}$ as the set of propositions which can be formed by disjunctions of the $m := 2^k$ different possible worlds. Suppose the uncertainty about the true possible world is described by a probability measure $p : \mathcal{F} \to [0, 1]$ which assigns a probability value $p_\tau := P(W_\tau)$ to each $W_\tau$. This measure is completely defined by the parameter vector $p := (p_1, \ldots, p_m)$. Let $\mathcal{Q} := \{(p_1, \ldots, p_m) \mid p_i \geq 0, \sum_{i=1}^{m} p_i = 1\}$ be the set of all probability measures over $\mathcal{F}$.

Information on the probability measure (or its characteristics) may be available from different sources: physical measurement devices, frequency counts from samples, or expert judgements. Assume there are $d$ pieces of evidence $\tilde{\pi} := (\tilde{\pi}_1, \ldots, \tilde{\pi}_d)$ on $p$. Each $\tilde{\pi}_i$ may be related to any measurable function $\pi_i(p)$ of the parameters $p$. For example $\tilde{\pi}_i$ may be equal to a probability $p(A)$, a conditional probability $p(A \mid B)$, an odds ratio $p(A \mid B)/p(A \mid \neg B)$, or an indicator which states, for instance, whether $p(A \wedge B) < 0.1$ or $p(A) > p(B)$ is true. The different $\tilde{\pi}_i$ may be uncertain and incoherent to some extent and are considered as some sort of *measurement* with an associated measurement noise. For physical measurements the noise results from perturbations of the measurement process, while for statistical samples it is caused by sampling fluctuations. For expert judgements measurement noise can be used to model the large fluctuations and incoherence of probability values specified by experts [Hogart 1987].

An independent *investigator* has the task to integrate these measurements. For him the measurement values $\tilde{\pi}$ and the probabilities $p$ are part of the external world about which he is uncertain and whose random relation is captured by a joint probability distribution $P(\tilde{\pi}, p)$. The capital $P$ indicates that the distribution is a 'second order' probability distribution describing random variables which themselves are probabilities. We suppose that the investigator can specify a subjective *measurement distribution* $P(\tilde{\pi} \mid p)$ reflecting his subjectively assessed distribution of possible measurements $\tilde{\pi}$ for each probability vector $p \in \mathcal{Q}$. Here we assume that the investigator acts as a rational agent who is able to formulate his subjective opinion in the form of coherent probability distributions. Note that the investigator does not specify any preference for some specific $p$ as he defines $P(\tilde{\pi} \mid p)$ for every possible value of $p$.

The measurement distribution is selected according to the type of measurement, e.g. the normal distribution for continuous statistics or the binomial distribution for probabilities [Ginsberg 1985, Paass 1986]. The latter seems appropriate if the expert bases his judgement on the experience gained from a number of cases. Under specific circumstances the measurement distribution may be very simple. Assume an expert has to answer the question "Is $p(A) \geq 0.8$ ?". Then $\tilde{\pi}_i$ can take the two values 0 or 1 meaning 'yes' or 'no'. The noise distribution just consists of two probability values: the probability that the expert would say 'yes' if really $p(A) \geq 0.8$ and the probability that the expert would say 'yes' if $p(A) < 0.8$.

## 3 The Bayesian Posterior Distribution

The measurement distribution $P(\tilde{\pi} \mid p)$ is the investigator's subjective assessment of the relation between $p$ and the measurement data $\tilde{\pi}$. The marginal distribution $P(p)$ can be interpreted as the investigator's prior distribution which describes his subjective state of knowledge on $p$ before he knows any measurements. Hence the joint distribution $P(\tilde{\pi}, p) = P(\tilde{\pi} \mid p)P(p)$ entirely rests on his subjective judgements. Then *Bayes theorem* yields his subjective *posterior density* of $p$ after the information $\tilde{\pi}$ has been taken into account

$$P(p \mid \tilde{\pi}) = \frac{P(\tilde{\pi} \mid p)P(p)}{P(\tilde{\pi})} \qquad (1)$$

The likelihood $P(p \mid \tilde{\pi})$ assigns a value to each $p$ which characterizes the degree of belief of the investigator that $p$ is the true probability distribution. For given $\tilde{\pi}$ the denominator $P(\tilde{\pi})$ is a constant. Note that the measurements $\tilde{\pi}_i$ of the different sources may be correlated with each other. This happens, for example, if two measurements are based on some common information. However, the formula is simplified considerably if we assume that each $\tilde{\pi}_i$ has been determined *independently* from the others. Then the joint likelihood is a product

$$P(\tilde{\pi} \mid p) := \prod_{i=1}^{d} P(\tilde{\pi}_i \mid p) \qquad (2)$$

In the important case that we have a random sample of size $n^i$ and we only observe whether a proposition $B \in \mathcal{F}$ holds or not, we get the binomial likelihood

$$P(\tilde{\pi}_i \mid p) \qquad (3)$$
$$\propto \pi_i(p)^{n^i \tilde{\pi}_i} + (1 - \pi_i(p))^{n^i(1-\tilde{\pi}_i)}$$

If for some measurement $\tilde{\pi}$ there is no $p$ such that $P(\tilde{\pi} \mid p)$ is greater than zero the measurement is inconsistent and no posterior distribution exists. The investigator may avoid this problem, for instance by assigning a possibly low but nonzero likelihood $P(\tilde{\pi} \mid p) > 0$ to each pair $p, \tilde{\pi}$.

The posterior distribution $P(p \mid \tilde{\pi})$ is a common joint density of the possible worlds. In general, however, the investigator is interested in the distribution $P(f(p) \mid \tilde{\pi})$ of more global characteristics $f(p)$, for instance $f(p) := p(A)$ for some $A \in \mathcal{F}$. To determine the posterior probability that $f(p)$ is in a specific interval, say $[a, b]$ he has to determine

$$P(M_{[a,b]} \mid \tilde{\pi}) = \int_{M_{[a,b]}} P(p \mid \tilde{\pi}) dp \qquad (4)$$

where $M_{[a,b]} := \{p \in \mathcal{Q} \mid f(p) \in [a, b]\}$. Except for trivial cases the analytical solution is intractable and we have to use a numerical approach. Conceptually we generate a sequence of $q_i \in \mathcal{Q}$ distributed according to $P(q \mid \tilde{\pi})$ and approximate the integral by $\frac{1}{\tau} \sum_{i=1}^{\tau} F(q_i)$ where $F(p)$ is the indicator function of $M_{[a,b]}$. If the $q_i$ are the realizations of an ergodic stationary Markov chain with distribution $P(q \mid \tilde{\pi})$ the sum converges to the integral by the ergodic theorem.

## 4 Stochastic Generation of Samples

The vector $p$ has $m = 2^k$ elements; a number which already for a moderate number $k$ of basic propositions is prohibitively large. Therefore we approximate $p$ by a random sample of $n$ possible worlds $W_\tau$. Let $\mathcal{Q}_n \subset \mathcal{Q}$ be the set of probability vectors with values in $\{\frac{0}{n}, \frac{1}{n}, \ldots, \frac{n}{n}\}$. Then at most $n$ different probabilities are larger than zero. According to the Law of Large Numbers any distribution $p \in \mathcal{P}$ can be approximated arbitrary well by a sample $q \in \mathcal{Q}_n$ if the sample size $n$ is chosen sufficiently large.

To generate the ergodic sequence $q \in \mathcal{Q}_n$ we utilize the Metropolis algorithm [Kalos & Whitlock 1986, p.73ff]. Let $X_\tau := (W_{\tau(1)}, \ldots, W_{\tau(n)})$ be a sample containing $n$ of the $m$ possible worlds. The algorithm starts with an arbitrary sample. In an iterative fashion the 'current' sample $X_\tau$ is modified to a new sample $X_\eta$ and subsequently it is checked whether the modification can be accepted. A modification usually consist of rather small changes, for instance transforming $U_i$ to $\neg U_i$ in one or more possible worlds $W_\tau(j)$. The probability $P_{mod}(X_\tau, X_\eta)$ of modifying $X_\tau$ to $X_\eta$ may be derived from some real, nonnegative, and

symmetric function $g(X_\tau, X_\eta)$

$$P_{mod}(X_\tau, X_\eta) = \frac{g(X_\tau, X_\eta)}{g(X_\tau)} \quad (5)$$

$$g(X_\tau) := \sum_\eta g(X_\tau, X_\eta) \quad (6)$$

A trivial choice is $P_{mod}(X_\tau, X_\eta) = P_{mod}(X_\eta, X_\tau) \geq 0$. If $q(X_\tau) \in \mathcal{Q}_n$ is the empirical distribution corresponding to $X_\tau$ the modification is accepted with probability

$$P_{acc}(X_\tau, X_\eta) := \max\left[1, \frac{P(\tilde{\pi} \mid q(X_\eta))}{P(\tilde{\pi} \mid q(X_\tau))}\right] \quad (7)$$

If each $X_\tau$ can be transformed into any other $X_\eta$ by a finite number of modifications, the probability $\Pr(X \mid \tilde{\pi})$ of $X$ being generated converges to an unique stationary distribution

$$\Pr(X \mid \tilde{\pi}) = c_1 g(X) P(\tilde{\pi} \mid q(X)) \quad (8)$$

as the number of iterations goes to infinity [Mitra & al. 1986]. Here $c_1$ is a constant normalizing the sum of probabilities to one. Hence the corresponding stationary distribution of the $p \in \mathcal{Q}_n$ is proportional to the posterior distribution

$$\begin{aligned}\Pr(p \mid \tilde{\pi}) &= \sum_{X: q(X)=p} P(X \mid \tilde{\pi}) \\ &= P(\tilde{\pi} \mid p) g(p) \quad (9)\end{aligned}$$

with $g(p) := \sum_{X: q(X)=p} c_1 g(X)$. In the simple case of symmetric modification probabilities, i.e. $P_{mod}(X_\eta, X_\tau) = P_{mod}(X_\tau, X_\eta)$ for all $X_\eta, X_\tau$, the term $g(p)$ is a constant.

If we use $P(\tilde{\pi} \mid q(X))^\beta$ instead of $P(\tilde{\pi} \mid q(X))$ in (7) and let $\beta$ grow to infinity, then according to simulated annealing theory [Mitra & al. 1986] the corresponding stationary distribution $\Pr_\beta(X \mid \tilde{\pi})$ concentrates on the set of maximum values of the aposteriori distribution $P(X \mid \tilde{\pi})$. For non-informative priors the resulting solution is equal to the maximum likelihood solution. But how can we interpret the term $P(\tilde{\pi} \mid q(X))^\beta$? In the joint likelihood $P(\tilde{\pi} \mid p) = \prod_i P(\tilde{\pi}_i \mid p)$ each individual term $P(\tilde{\pi}_i \mid p)$ represents $n_i \geq 1$ observations of the statistic $\tilde{\pi}_i$. If we had two independent samples with the same observations we would get the term $P(\tilde{\pi}_i \mid p) * P(\tilde{\pi}_i \mid p)$ in the joint likelihood. Hence a joint likelihood $P(\tilde{\pi} \mid q(X))^\beta$ corresponds to the situation that for each individual term we observe the same statistic $\tilde{\pi}_i$ from $\beta * n_i$ instead of $n_i$ independent observations.

Starting with an arbitrary sample $X$ the updating scheme (6) and (7) eventually will converge to the stationary distribution (8). However the convergence to this distribution may take a long time if the constraints imposed by the data $\tilde{\pi}$ are difficult to accomplish. To speed up the convergence we may start the procedure with a value $\beta \ll 1$ and gradually increase $\beta$ to 1. Simulated annealing theory shows that in this way the desired stationary distribution is approached much faster than in the case that we fix $\beta$ to 1.

## 5  Prior Distributions

To utilize the Bayesian approach the investigator has to select a prior probability distribution $P(p)$. If he has no preference for some ranges of probability he can specify a *noninformative prior* which favours no $p$ over others. Because of its invariance to transformations, textbooks [Berger 1980 p.74ff] recommend versions of the Dirichlet density which is proportional to $\prod_\tau p(W_\tau)^{\alpha-1}$, usually with $\alpha < 1$. There is, however, still a debate on which prior probability to choose [Hartigan 1983 p.96].

In general the available evidence concerns only a few lower order characteristics of $p$. The distribution of all higher order interactions is completely determined by the prior distribution. Hence in some respect the choice of a noninformative prior makes explicit the structural hypotheses which in other updating formalisms are hidden in the maximum entropy assumption [cf. Cheeseman 1985]. As, however, prior distributions usually give nonzero density to *every* parameter value, they are far less restrictive than the maximum entropy assumption. In our algorithm a prior density can be integrated in two different ways: we first may simply use the product $P(\tilde{\pi} \mid q(X)) P(q(X))$ of the prior density and the likelihood to determine the probability of acceptance in (7). Obviously we may incorporate arbitrary priors in this way. Alternatively we may define an appropriate $g(X_\tau, X_\eta)$ such that the prior distribution evolves as the stationary distribution of the modification process (6).

In our numerical experiments we used a uniform prior for each $p(U_i)$, which is a special case of the Dirichlet distribution. It was generated by defining an appropriate $g(X_\tau, X_\eta)$. Assume $p(U_i)$ has a specific value, e.g. $p(U_i) = 0.7$. Then with equal probability $p(U_i)$ is increased or reduced for a small amount $\delta$. In the case of increase, for instance, we randomly select one or more possible worlds where $\neg U_i$ holds from the current sample $X_\tau$ and change them to $U_i$. Hence $P(U_i)$ follows a random walk with the reflecting barriers 0 and 1. As the mean distance to the starting value of $p(U_i)$ after $k$ modifications is proportional to $\delta k^{1/2}$, we have to select $\delta$ large enough, to arrive at the uniform distribution of $p(U_i)$ in a short enough time.

The modifications of the different $p(U_i)$ were performed mutually independent. From the independence follows that for $\tilde{U}_i \in \{U_i, \neg U_i\}$ the *conditional probability* $p(U_i \mid \tilde{U}_{j_1} \wedge \cdots \wedge \tilde{U}_{j_l})$ for arbitrary $j_r \neq i$ has a *uniform* prior distribution too. If the resulting posterior distribution $P(p(U_i \mid U_j) \mid \tilde{\pi})$ of some conditional probability $p(U_i \mid U_j)$ is different from the uniform distribution, this effect is caused by the observed data. This seems to be a desirable property for the evaluation of inference networks. The $g(X_\tau, X_\eta)$-terms resulting from this scheme of generating the prior distribution correspond to the conditions ( 6) and hence the stationary distribution (8) results.

If we want to determine the probability of some $B$ for a situation where $U_i$ is known to hold, we have to estimate the conditional probability $p(B \mid U_i)$. Whenever the probability of $U_i$ is low then a sample $X_\tau$ usually will comprise only few or none possible worlds $W_\eta$ where $U_i$ holds and the estimated posterior distribution for $p(B \mid U_i)$ will exhibit a high variance. This situation may be avoided by using *weighted samples* to allow that possible worlds with low weights may be generated where $U_i$ holds. To each possible world $W_{\tau(j)}$ in the sample $X_\tau := (W_{\tau(1)}, \ldots, W_{\tau(n)})$ a weight $w_j \geq 0$ is attached with $\sum_i w_i = 1$. For half of the possible worlds the $i$-th basic proposition is fixed to $U_i$ while for the rest it is set to $\neg U_i$. The prior distribution for $U_i$ is generated by changing the weights such that $p(U_i)$ follows a random walk with reflecting barriers 0 and 1.

All definitions given above may be easily extended to the case that the basic aspects of the situation are characterized by more than two mutually exclusive and exhaustive propositions $U_{i1}, \ldots, U_{ik_i}$. Again we can use independent Dirichlet priors for the marginal distribution of the different aspects. If an aspect corresponds to a numerical variable $u$, we can either model it by a large number of mutually exclusive propositions $U_{i1}, \ldots, U_{ik_i}$ with each $U_{il}$ corresponding to some value $\xi_{il}$ of $u_i$. Alternatively one can include a continuous $u_i$ directly without any discretization using the Markov chain results of [Kalos & Whitlock 1986]. Then we could take into account measurements which, for instance, state that the mean value of $u_i$ is equal to 13.6 or that $u_i \geq 10.5$ with probability 0.8.

## 6  Dempster-Shafer Theory

As shown by [Kyburg 1987, Paass 1988] every Dempster-Shafer belief function may be expressed by inequality constraints on an underlying probability measure $p$. In the framework of this paper we may consider a probability mass function as the result of a measurement process. This shall be demonstrated by a small example. Assume there are $m = 3$ possible worlds $W_1, W_2, W_3$ and let $\mathcal{F}$ be the corresponding Boolean algebra with 8 elements $\mathcal{F} := \{\emptyset, W_1, \ldots, W_3, W_1 \vee W_2, \ldots, W_1 \vee W_2 \vee W_3\}$. Assume information on $p$ is contained in the following probability mass function $\mu : \mathcal{F} \to [0,1]$:

$\mu(W_1) = 0.3;\quad\quad\quad \mu(W_3) = 0.1;$
$\mu(W_2 \vee W_3) = 0.4;\quad \mu(W_1 \vee W_2 \vee W_3) = 0.2;$

Let us define an observable variable $\pi$ taking the non-negative values $\tilde{\pi}_1 = \mu(W_1)$, $\tilde{\pi}_3 = \mu(W_3)$, $\tilde{\pi}_{23} = \mu(W_2 \vee W_3)$, $\tilde{\pi}_{123} = \mu(W_1 \vee W_2 \vee W_3)$ summing up to 1. Then we may define a mapping $\tilde{\pi} = H(\alpha)p$ by

$$\begin{pmatrix} \tilde{\pi}_1 \\ \tilde{\pi}_3 \\ \tilde{\pi}_{23} \\ \tilde{\pi}_{123} \end{pmatrix} = \begin{pmatrix} \alpha_{11} & 0 & 0 \\ 0 & 0 & \alpha_{31} \\ 0 & \alpha_{21} & \alpha_{32} \\ 1-\alpha_{11} & 1-\alpha_{21} & 1-\alpha_{31}-\alpha_{32} \end{pmatrix} \begin{pmatrix} p(W_1) \\ p(W_2) \\ p(W_3) \end{pmatrix}$$

with unknown parameters $\alpha_{ij}$. Now we may de-fine $P(\tilde{\pi} \mid p) = const > 0$ if $\tilde{\pi} = H(\alpha)p$ such that all $\alpha_{ij} \in [0,1]$ and $P(\tilde{\pi} \mid p) = 0$ otherwise. After proper normalization $P(\tilde{\pi} \mid p)$ may be considered as a measurement distribution. Using the Bayesian Formula (1) we may derive a posterior distribution $P(p \mid \tilde{\pi})$. If in addition we use a non-informative prior $P(p) = const$, the posterior $P(p \mid \tilde{\pi})$ is constant on its support $M_{\tilde{\pi}} := \{p \subseteq \mathcal{Q} \mid \exists_\alpha \tilde{\pi} = H(\alpha)p\}$. From this set upper and lower probabilities for arbitrary $B \in \mathcal{F}$ may be determined.

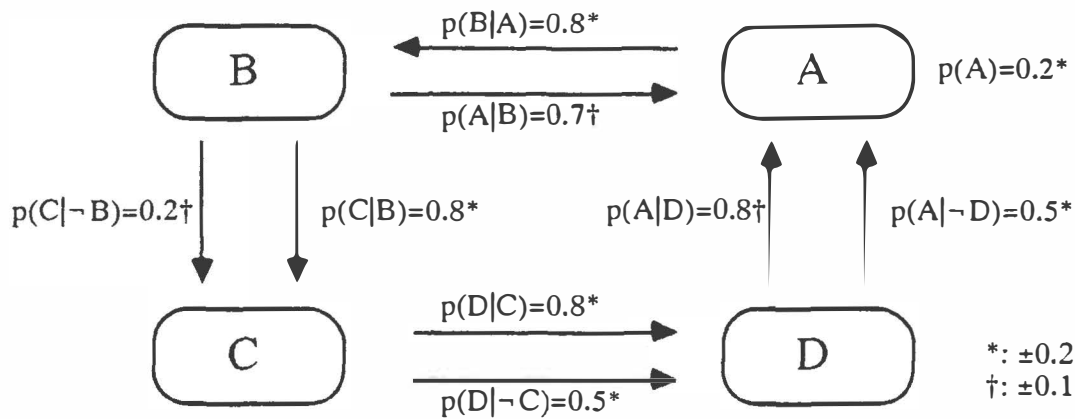

Figure 1: **Inference Network for Numerical Demonstration**

The measurement distribution may be enhanced in the usual way to cover situations where the assignment of a sample element to some $\tilde{\pi}_i$ may be subject to error. More important we may have different samples corresponding to different probability mass functions. Then the Bayesian Formula (1) is a way to to combine these different pieces of evidence with the sample size indicating their reliability.

## 7 Numerical Demonstration

We applied the algorithm to a small inference network with uncertain rules which form a cycle. Inference nets with this general structure cannot be handled directly by other inference techniques requiring an ordering of the nodes. In the model two 'symptoms' $A$ and $B$ affect the probability of the 'outcomes' $C$ and $D$. We assumed that the probabilities given in figure (1) had been estimated by independent experts. The investigator assigned a binomial measurement distribution to the experts estimating that the values supplied were exact to $\pm 0.1$ or $\pm 0.2$ in 90% of the cases. These intervals were used to derive the parameters of the corresponding binomial measurement distributions giving likelihoods (4).

For the simulation of the posterior distribution $P(p \mid \tilde{\pi})$ a sample $X$ of size $n = 200$ was utilized. In a first analysis the information $p(A \mid B) = 0.7$ was omitted. Using the Metropolis algorithm with $\beta = 1$ the distribution of the sample quickly reached a stationary state. The characteristics of the posterior distribution of each possible world were determined according to (4) and are shown on the left side of table (1). The probability estimates show a considerable variance around the the median, the 50%-percentile of the posterior distribution. The difference between the 25%-percentile and the 75%-percentile in many rows is larger than the median.

The probabilities derived so far describe the general stochastic relation between the propositions. Now assume that evidence arrives that for a specific case one or more of the 'symptoms' $B$ and $A$ hold. The resulting probability is given by the posterior distribution of the corresponding conditional probability, e.g. $p(D \mid B)$. Three such distributions were determined using (4) and are given in the last lines of table (1). Again the distributions have a rather large spread. The observation of the new fact $\neg B$ not always leads to a reduction of variance as shown by the comparison of $p(D \mid \neg B)$ and $p(D \mid \neg A \wedge \neg B)$.

Table 1: **Percentiles of the Posterior Distributions for Selected Probabilities**

|  | Percentiles ||||||||||
|---|---|---|---|---|---|---|---|---|---|---|
|  | $p(A \mid B) = 0.7$ omitted ||||| $p(A \mid B) = 0.7$ included |||||
|  | 10% | 25% | 50% | 75% | 90% | 10% | 25% | 50% | 75% | 90% |
| $p(\neg A \neg B \neg C \neg D)$ | .012 | .025 | **.042** | .062 | .073 | .060 | .090 | **.128** | .150 | .157 |
| $p(\neg A \neg B \neg C D)$ | .002 | .007 | **.014** | .030 | .043 | .002 | .003 | **.010** | .021 | .032 |
| $p(\neg A \neg B C \neg D)$ | .000 | .000 | **.003** | .010 | .017 | .000 | .002 | **.003** | .011 | .025 |
| $p(\neg A \neg B C D)$ | .000 | .000 | **.008** | .008 | .017 | .000 | .000 | **.008** | .008 | .017 |
| $p(\neg A B \neg C \neg D)$ | .002 | .009 | **.019** | .030 | .038 | .002 | .005 | **.010** | .030 | .053 |
| $p(\neg A B \neg C D)$ | .008 | .017 | **.025** | .037 | .053 | .008 | .017 | **.025** | .042 | .050 |
| $p(\neg A B C \neg D)$ | .008 | .017 | **.033** | .058 | .091 | .000 | .008 | **.025** | .038 | .050 |
| $p(\neg A B C D)$ | .050 | .067 | **.092** | .123 | .166 | .050 | .067 | **.075** | .099 | .124 |
| $p(A \neg B \neg C \neg D)$ | .030 | .043 | **.055** | .073 | .085 | .000 | .003 | **.010** | .026 | .053 |
| $p(A \neg B \neg C D)$ | .009 | .024 | **.040** | .066 | .108 | .002 | .008 | **.017** | .033 | .066 |
| $p(A \neg B C \neg D)$ | .000 | .000 | **.008** | .017 | .026 | .000 | .002 | **.008** | .017 | .028 |
| $p(A \neg B C D)$ | .008 | .017 | **.033** | .042 | .068 | .017 | .017 | **.033** | .050 | .066 |
| $p(A B \neg C \neg D)$ | .010 | .017 | **.027** | .051 | .071 | .008 | .017 | **.027** | .047 | .058 |
| $p(A B \neg C D)$ | .050 | .067 | **.091** | .117 | .153 | .051 | .075 | **.100** | .125 | .149 |
| $p(A B C \neg D)$ | .033 | .050 | **.075** | .117 | .143 | .059 | .075 | **.108** | .156 | .192 |
| $p(A B C D)$ | .258 | .317 | **.375** | .433 | .491 | .225 | .294 | **.358** | .408 | .467 |
| $p(D \mid B)$ | .589 | .694 | **.781** | .841 | .906 | .550 | .675 | **.765** | .813 | .845 |
| $p(D \mid \neg B)$ | .259 | .360 | **.486** | .574 | .629 | .183 | .245 | **.330** | .390 | .498 |
| $p(D \mid \neg A \land \neg B)$ | .098 | .221 | **.347** | .441 | .593 | .019 | .064 | **.120** | .209 | .339 |

Assume an additional expert states $p(A \mid B) = 0.70$ and the investigator judges this measurement to be rather reliable with a 90%-interval of $\pm 0.1$. The right part of table 1 contains the resulting posterior distributions. While the distribution of $p(D \mid B)$ is only slightly affected by the modification, the medians of $p(D \mid \neg B)$ and $p(D \mid \neg A \land \neg B)$ are sharply reduced. If in addition $p(B \mid \neg A) = 0.3$ is introduced with a 90%-interval of $\pm 0.1$ this information is contradictory to some extend. By comparing the resulting mean posterior probabilities with the initial measurement distributions we can assess the extend of contradiction for each piece of information. The observation $p(A \mid B) = 0.7$ turns out to be 'most contradictory' in this sense. This demonstrates how incompatible pieces of evidence can be spotted during the evaluation of the inference network.

In a last test we applied the algorithm to a larger randomly generated inference network with 40 basic propositions. There were more than 400 uncertain restrictions on probabilities relating up to four randomly selected basic propositions. The inference network contained many loops and cycles. The measurement distribution was chosen to be multinomial. After 2 minutes processing time on an IBM 3093 the algorithm had reached stationary state. This experiment shows that the procedure may be applied to larger inference networks with arbitrary structure. The result is plausible as the underlying simulated annealing algorithm is a global optimization procedure with good convergence properties.

## 8 Summary and Discussion

We have presented an algorithm that is able to integrate uncertain probability estimates and to approximate the corresponding Bayesian second order posterior distribution by the simulation of a random sample. The approach is applicable to inference networks of arbitrary structure. The approximation of the distribution by a sample allows to capture the basic stochastic relations while being numerically feasible for larger networks. The resulting posterior distribution reflects the uncertainty in probability estimates and can directly be used for decision purposes. The stochastic simulation algorithm is part of the simulated annealing algorithm which nor-mally is employed to solve large constraint satisfaction problems. It is inherently parallel and can be implemented on parallel hardware.

In the light of the psychological studies cited above the uncertainty of the input probabilities of ±0.1 to ±0.2 seems to be plausible for realistic applications. The numerical experiments show that the variability of the results is of the same magnitude. This aspect is usually neglected by other methods for Bayesian networks which produce point estimates that may lead a user to risky decisions as he does not know the variability of the estimates. In addition to being in general not justified by the quality of input data, [Cooper 1990] has shown that computing the exact solution of Bayesian networks is NP-hard, and therefore proposes the utilization of approximate algorithms for Bayesian networks of general type. The above algorithm is approximate and derives the variability of the results. By increasing the sample size $n$, the accuracy may be adapted such that it meets the variability of the results, is sufficient for the desired purpose, and remains computationally feasible.

An important issue, which remains to be studied in more detail, concerns the properties of the different 'noninformative' prior distributions. This amounts to specifying the state of complete ignorance about the unknown probability distribution. The prior used in this paper has the advantage of uniform marginal and conditional densities, while other priors are insensitive to transformations (Dirichlet priors) or concentrate on maximum entropy configurations [Berger 1980 p.75]. Especially the latter approach seems to be promising for future investigations.